# Extended depth-of-field in holographic image reconstruction using deep learning based auto-focusing and phase-recovery

**YICHEN WU,**[1,2,3,†] **YAIR RIVENSON,**[1,2,3,†] **YIBO ZHANG,**[1,2,3] **ZHENSONG WEI,**[1] **HARUN GÜNAYDIN,**[1] **XING LIN,**[1,2,3] **AYDOGAN OZCAN** [1,2,3,4,*]

[1]*Electrical and Computer Engineering Department, University of California, Los Angeles, California 90095, USA*
[1]*Bioengineering Department, University of California, Los Angeles, California 90095, USA*
[1]*California NanoSystems Institute (CNSI), University of California, Los Angeles, California 90095, USA*
[1]*Department of Surgery, David Geffen School of Medicine, University of California, Los Angeles, California 90095, USA*
[†]*Equal contribution authors*
*Corresponding author: ozcan@ucla.edu

**Holography encodes the three dimensional (3D) information of a sample in the form of an intensity-only recording. However, to decode the original sample image from its hologram(s), auto-focusing and phase-recovery are needed, which are in general cumbersome and time-consuming to digitally perform. Here we demonstrate a convolutional neural network (CNN) based approach that simultaneously performs auto-focusing and phase-recovery to significantly extend the depth-of-field (DOF) in holographic image reconstruction. For this, a CNN is trained by using pairs of randomly de-focused back-propagated holograms and their corresponding in-focus phase-recovered images. After this training phase, the CNN takes a single back-propagated hologram of a 3D sample as input to rapidly achieve phase-recovery and reconstruct an in focus image of the sample over a significantly extended DOF. This deep learning based DOF extension method is non-iterative, and significantly improves the algorithm time-complexity of holographic image reconstruction from O(*nm*) to O(1), where *n* refers to the number of individual object points or particles within the sample volume, and *m* represents the focusing search space within which each object point or particle needs to be individually focused. These results highlight some of the unique opportunities created by data-enabled statistical image reconstruction methods powered by machine learning, and we believe that the presented approach can be broadly applicable to computationally extend the DOF of other imaging modalities.**

## 1. Introduction

Holography [1–6] encodes the three-dimensional (3D) information of a sample through interference of the object's scattered light with a reference wave. Through this interference process, the intensity of a hologram that is recorded by e.g., an image sensor, contains both the amplitude and phase information of the sample [7]. Retrieval of this object information over a 3D sample space has been the subject of numerous holographic imaging techniques [2–4,7–10]. In a holographic image reconstruction process, there are two major steps. One of these is the phase-recovery, which is required since only the intensity information of the holographic pattern is recorded at a given digital hologram. In general, for an off-axis holographic imaging system [3,4,9,10], this phase-recovery step can be achieved relatively easier compared to an in-line holography set-up, at the cost of a reduction in the space-bandwidth product of the imaging system. For in-line holography, on the other hand, iterative phase-recovery approaches that utilize measurement diversity and/or prior information regarding the sample have been developed [5,11–18]. Regardless of the specific holographic set-up that is employed, phase-recovery needs to be performed to get rid of the twin-image and self-interference related spatial artifacts in the reconstructed phase and amplitude images of the sample.

The other crucial step in holographic image reconstruction is auto-focusing, where the sample-to-sensor distances (i.e., relative heights) of different parts of the 3D object need to be numerically estimated. Auto-focusing accuracy is vital to the quality of the reconstructed holographic image such that the phase-recovered optical field can be back-propagated to the correct object locations in 3D. Conventionally, to perform auto-focusing, the hologram is digitally propagated to a set of axial distances, where a focusing criterion is evaluated at each resulting complex-valued image. This step is ideally performed after the phase-recovery step, but can also be applied before it, which might reduce the focusing accuracy [19]. Various auto-focusing criteria have been successfully used in holographic imaging, including e.g., the Tamura coefficient [20], the Gini Index [21] and others [19,22–27]. Regardless of the specific focusing criterion that is used, and even with smart search strategies [28], the auto-focusing step requires numerical back-propagation of optical fields and evaluation of a criterion at typically >10-20 axial distances, which is time-consuming for even a small field-of-view (FOV). Furthermore, if the sample has multiple objects at different depths, this procedure needs to repeat for every object in the FOV.

Some recent work has utilized deep learning to achieve auto-focusing. Z. Ren *et al.* formulated auto-focusing as a classification problem and used a convolutional neural network (CNN) to provide rough estimates of the focusing distance with each classification class (i.e., bin) having an axial range of ~3 mm, which is more appropriate for imaging systems that do not need precise knowledge of the axial distance of each object [29]. As another example, T. Shimobaba *et al.* used a CNN regression model to achieve continuous auto-focusing, also with a relatively coarse focusing accuracy of >5 mm [30]. In parallel to these recent results, CNN-based phase-recovery methods that use a single intensity-only hologram to reconstruct a two-dimensional object's image have also been demonstrated [31–34]. However, in these former approaches the neural networks were trained with *in-focus* images, where the sample-to-sensor (hologram) distances were precisely known *a priori* based on the imaging set-up or were separately determined based on an auto-focusing criterion. As a result, the reconstruction quality degraded rapidly outside the system depth-of-field (DOF); for example, for high resolution imaging of a pathology slide (tissue section), ~4 μm deviation from the correct focus distance resulted in loss of resolution and distorted the sub-cellular structural details. [34]

Here, we demonstrate a deep learning based holographic image reconstruction method that performs both auto-focusing and phase-recovery at the same time using a single hologram intensity, which significantly extends the DOF of the reconstructed image compared to previous approaches, while also improving the algorithm time-complexity of holographic image reconstruction from O($nm$) to O(1). We term this approach as *HIDEF* (Holographic Imaging using Deep learning for Extended Focus) and it relies on training a CNN with not only in-focus image patches, but also with randomly de-focused holographic images along with their corresponding in-focus and phase-recovered images, used as reference. Overall, HIDEF boosts the computational efficiency of high-resolution holographic imaging by simultaneously performing auto-focusing and phase-recovery and increases the robustness of the image reconstruction process to potential misalignments in the optical set-up by significantly extending the DOF of the reconstructed images. In addition to digital holography, the same deep learning-based approach can also be applied to improve the DOF of incoherent imaging modalities including e.g., fluorescence microscopy. This work and its results highlight some of the exciting opportunities created by deep learning-based statistical image reconstruction approaches that provide unique solutions to challenging imaging problems, enabled by the availability of high-quality image data.

## 2. Methods

The architecture of the neural network that we used is shown in Fig. 1. This CNN architecture is inspired by U-Net [35], and it consists of a down-sampling path as well as a symmetric up-sampling path, each including convolution blocks, with a kernel size of 3×3, followed by Rectified Linear Unit (ReLU) operators. Through a chain of down-sampling operations, the network learns to capture and separate the true image and twin image spatial features of a holographic input field at different scales [36]. Additional short-cut paths (blue arrows in Fig. 1) are also included to pass the information forward through residual connections, which are useful to increase the training speed of the network [37]. This CNN architecture is implemented using TensorFlow, an open-source deep learning software package [38]. During the training phase, the CNN minimizes the l1-norm distance of the network output from the target/reference images, and iteratively updates the network's weights and biases using the adaptive moment estimation (Adam) optimizer [39], with a learning rate of $10^{-4}$. For each image dataset, the ratio of the training to cross-validation was set to 14:3. The training and blind testing of the network were performed on a PC with six-core 3.60 GHz CPU, 16GB of RAM, using Nvidia GeForce GTX 1080Ti GPU. On average, the training process takes ~40 h for e.g., 200,000 iterations, corresponding to ~100 epochs. After the training, the network inference time for a hologram patch of 512 x 512 pixels (with phase and amplitude channels) is < 0.3 s.

## 3. Results and Discussion

To demonstrate the success of HIDEF, in our initial set of experiments, we used aerosols that are captured by a soft impactor surface and imaged by an on-chip holographic microscope, where the optical field scattered by each aerosol interferes with the directly transmitted light forming an in-line hologram, sampled using a CMOS imager, without the use of any lenses [28]. The captured aerosols on the substrate are dispersed in multiple depths ($z_2$) as a result of varying particle mass, flow speed, and flow direction during the air sampling period [28]. Based on this set-up, the training image dataset had 176 digitally-cropped non-overlapping regions that only contained *particles located at the same depth*, which are further augmented by 4-fold to 704 regions by rotating them to 0, 90, 180 and 270 degrees. For each region, we used a single hologram intensity and back-propagated it to 81 random distances, spanning an axial range of -100 μm to 100 μm away from the correct global focus, determined by auto-focusing using the Tamura of the Gradient criterion [19]. We then used these complex-valued fields as the input to the network. The target images used in the training phase (i.e., the reference images corresponding to the same samples) were reconstructed using multi-height phase-recovery (MH-PR) that utilized 8 different in-line holograms of the sample, captured at different $z_2$ distances, to iteratively recover the phase information of the sample, after an initial auto-focusing step performed for each height [40].

After this training phase, next we blindly tested the HIDEF network on samples that had *no overlap* with the training or validation sets; these samples contained *particles spread across different depths per image FOV*. Fig. 2 illustrates the success of HIDEF and how it simultaneously achieves an extended DOF and phase-recovery. For a given in-line hologram of the captured aerosols (Fig. 2(a)), we first back-propagate the hologram intensity to a coarse distance of $z_2$ = 1 mm away from the active area of the CMOS imager, which is roughly determined based on the effective substrate thickness used in the experiment. This initial back-propagated hologram yields a strong twin image because of the short propagation distance (~1 mm) and the missing phase information. This complex-valued field, containing both the true and twin images, is then fed to the CNN. The output of the CNN is shown in Fig. 2(a), which demonstrates the extended DOF of HIDEF with various aerosols, spread over an axial range of ~90 μm, that are all brought into focus at the network output. In addition to bringing all the particles contained in a single hologram to a sharp focus, the network also performed phase-recovery, resulting in phase and amplitude images that are free from twin image and self-interference related artifacts. Fig. 2(b) and (c) also compare the results of the network output with respect to a standard MH-PR approach that used eight in-line holograms to iteratively retrieve the phase information of the sample. These comparisons clearly demonstrate both the significantly extended DOF and phase-recovery performance of HIDEF, achieved using a single hologram intensity with a non-iterative inference time of < 0.3 s. In comparison, the iterative MH-PR approach took ~4 s for phase-recovery and an additional ~2.4 s for auto-focusing to the individual objects at eight planes, totaling ~6.4 s for the same FOV and object volume, i.e., >20-fold slower compared to HIDEF.

In these results, we used a coarse back-propagation step of 1 mm, before feeding the CNN with a complex-valued field. An important feature of our approach is that this back-propagation distance, $z_2$, does not need to be precise. In Visualization 1, we demonstrate the stability of the HIDEF output image as we vary the initial back-propagation distance, providing the same extended DOF image regardless of the initial $z_2$ selection. This is very much expected since the network was trained with defocused holograms spanning an axial defocus (dz) range of +/- 0.1 mm. For this specific FOV shown in Visualization 1, all the aerosols that were randomly spread in 3D experienced a defocus amount that is limited by +/- 0.1 mm (with respect to their correct axial distance in the sample volume). Beyond this range of defocusing, the HIDEF network cannot perform reliable image reconstruction since it was not trained for that (see e.g., |dz| > 120 μm in Fig. 2(c)). In fact, outside of its training range, HIDEF starts to hallucinate features as illustrated in Visualization 2, which covers a much larger axial defocus range of -1 mm ≤ dz ≤ 1 mm – beyond what the network was trained for.

Interestingly, although the network was only trained with *globally* de-focused hologram patches that only contain particles at the same depth/plane, it learned to individually focus various particles that lie at different depths within the same FOV (see Fig. 2(a)). Based on this observation, one can argue that the HIDEF network does *not* perform the physical equivalent of free-space back-propagation of a certain hologram FOV to a focus plane. Instead, it statistically learns both in-focus and out-of-focus features of the input field, segments the out-of-focus parts and replaces them with in-focus features, in a parallel manner for a given hologram FOV. From an algorithm time-complexity perspective, this is a fixed processing time for a given hologram patch, i.e., a complexity of $O(1)$, instead of the conventional $O(nm)$, where $n$ defines the number of individual object points or particles within the 3D sample volume, and $m$ is the discrete focusing search space.

Based on the above argument, if the network statistically learns both in-focus and out-of-focus features of the sample, one could think that this approach should be limited to relatively sparse objects (such as the one shown in Fig. 2) that let the network learn out-of-focus sample features within a certain axial defocusing range, used in the training. In fact, to test this hypothesis with non-sparse samples, next we tested HIDEF on the holograms of spatially connected objects such as tissue slices, where there is no opening or empty region within the sample plane. For this goal, based on the CNN architecture shown in Fig. 1, we trained the network with 1,119 hologram patches (corresponding to breast tissue sections used in histopathology), which were randomly propagated to 41 distances spanning an axial defocus range of -100 μm to 100 μm with respect to the focus plane. In this training phase, we used MH-PR images as our target/reference. Our blind testing results, after the training of the network, are summarized in Fig. 3 and Visualization 3, which clearly demonstrate that HIDEF can simultaneously perform both phase-recovery and auto-focusing for an arbitrary, non-sparse and connected sample. In Fig. 3, we also see that MH-PR images naturally exhibit a limited DOF: even at an axial defocus of ~5 μm, some of the fine features at the tissue level are distorted. With more axial de-focus, the MH-PR results show significant artificial ripples and loss of further details. HIDEF, on the other hand, is very robust to axial defocusing, and is capable of correctly focusing the entire image and its fine features, while also rejecting the twin image artifact at different de-focus distances, up to the range that it was trained for (±0.1 mm) – see Visualization 3.

However, as illustrated in Visualization 4 and Fig. 3, beyond its training range, HIDEF starts to hallucinate and create false features. A similar behavior is also observed in Visualization 2 for the aerosol images. There are several messages that one can take from these observations: the network does not learn or generalize a specific physical process such as wave propagation, or light interference; if it were to generalize such physical processes, one would not see sudden appearances of completely unrelated spatial features at the network output as one gradually goes outside the axial defocus range that it was trained for. For example, if one compares the network output within the training range and outside (see Visualization 3 and 4), one can clearly see that we do not see a physical smearing or diffraction-related smoothening effect as one continues to defocus in a range that the network was not trained for. In this defocus range that is "new" to the network, it still gives relatively sharp, but unrelated features, which indicate that it is not learning or generalizing physics of wave propagation or interference.

To further quantify the improvements made by HIDEF, next we compared the amplitude of the network output image against the MH-PR result at the correct focus of the tissue section, and used the structural similarity (SSIM) index for this comparison, defined as [41]:

$$\text{SSIM}(U_1, U_2) = \frac{(2\mu_1\mu_2 + C_1)(2\sigma_{1,2} + C_2)}{(\mu_1^2 + \mu_2^2 + C_1)(\sigma_1^2 + \sigma_2^2 + C_2)} \quad (1)$$

where $U_1$ is the image to be evaluated, and $U_2$ is the reference image, which in this case is the auto-focused MH-PR result using eight in-line holograms. $\mu_p$ and $\sigma_p$ are the mean and standard deviation for image $U_p$ (p =1,2), respectively. $\sigma_{1,2}$ is the cross-variance between the two images, and $C_1, C_2$ are stabilization constants used to prevent division by a small denominator. Based on these definitions, Fig. 4 shows the mean SSIM index calculated across an axial de-focus range of -100 μm to 100 μm, which was averaged across 180 different breast tissue FOVs that were blindly tested. Consistent with the qualitative comparison reported in Fig. 3 and Visualization 3, HIDEF outputs SSIM values that are significantly higher than the hologram intensities back-propagated to the exact focus distances, owing to the phase-recovery capability of the network. Furthermore, as shown in Fig. 4, compared to a CNN that is trained using only in-focus holograms (with exact $z_2$ values), HIDEF has a much higher SSIM index for de-focused holograms, across a large DOF of ~0.2 mm. Interestingly, the network that is trained with in-focus holograms beats HIDEF for only one point in Fig. 4, i.e., for dz = 0 μm, which is expected as this is what it was specifically trained for. However, this small difference in SSIM (0.78 vs. 0.76) is visually negligible (see Visualization 3, the frame of dz = 0 μm).

Our results reported so far demonstrate the unique capabilities of HIDEF to simultaneously perform phase-recovery and auto-focusing, yielding at least an order of magnitude increase in the DOF of the reconstructed images, as also confirmed by Figs. 2-3 and Visualizations 1-3. To further extend the DOF of the neural network output beyond 0.2 mm, one can use a larger network (with more layers, weights and biases) and/or more training data, containing severely defocused images as part of its learning phase. Certainly, the proof-of-concept DOF enhancement reported here is not an ultimate limit for the presented approach. In fact, to better emphasize this opportunity we also trained a third neural network, following the HIDEF architecture of Fig. 1, with a training image set that contained randomly defocused holograms of breast tissue sections, with an axial defocus range of -0.2 mm to 0.2 mm. The performance comparison of this new network against the previous one (demonstrated in Fig. 3) is reported in Fig. 4. As shown in this comparison, by using a training image set that included even more defocused holograms, we were able to significantly extend the axial defocus range to 0.4 mm (i.e., +/- 0.2 mm), where the HIDEF network successfully performed both auto-focusing and phase-recovery, at the same output image.

In summary, we believe that the results reported in this work provide compelling evidence for some of the unique opportunities created by statistical image reconstruction methods enabled by especially deep learning, and the approach presented here can be widely applicable to computationally extend the DOF of other imaging modalities, including e.g., fluorescence microscopy.

**Funding.** The Ozcan Research Group at UCLA acknowledges the support of NSF Engineering Research Center (ERC, PATHS-UP), the See Supplement for supporting content.

# Figures and Figure Captions

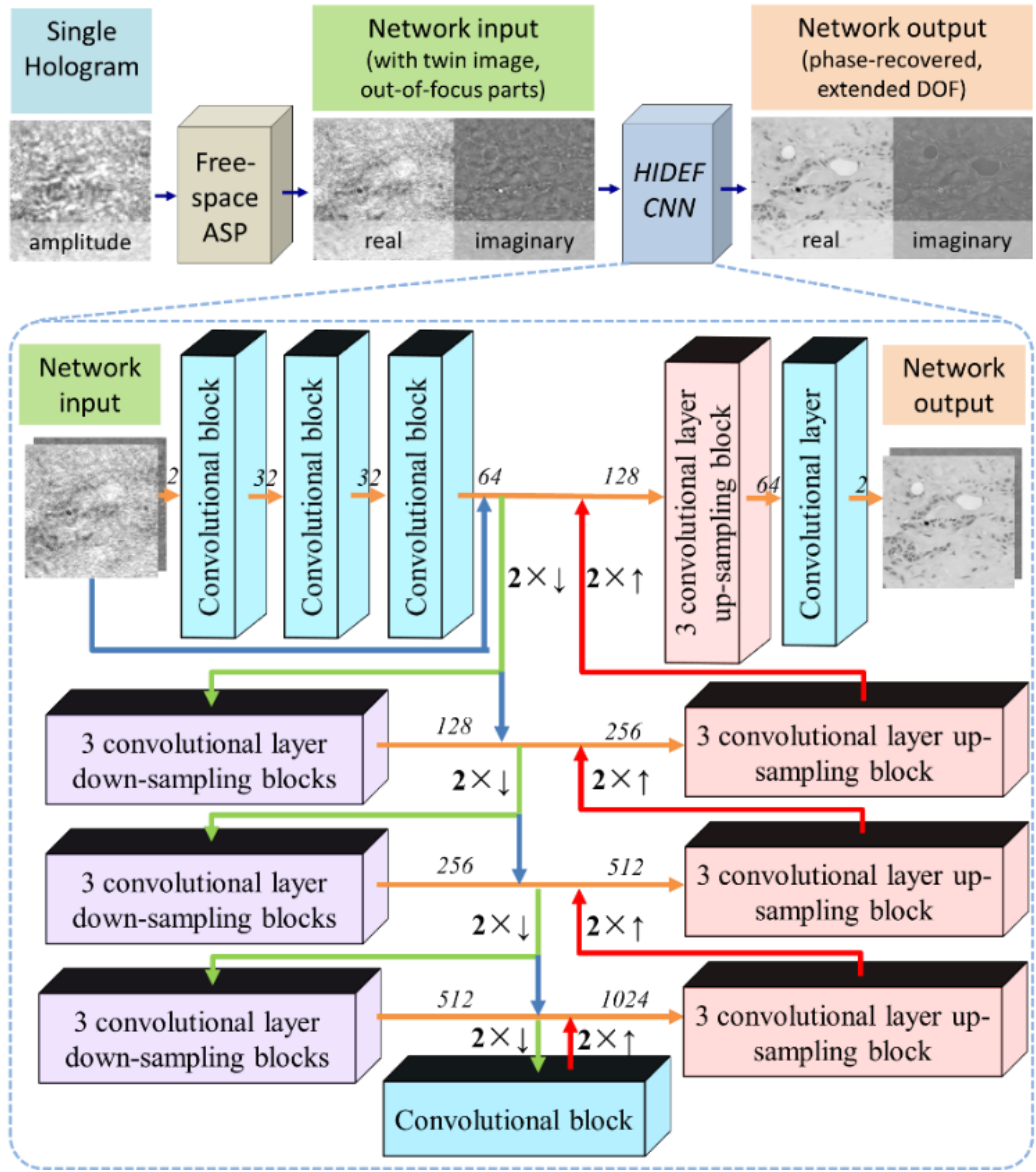

**Fig. 1.** HIDEF CNN, after its training, simultaneously achieves phase-recovery and auto-focusing, significantly extending the DOF of holographic image reconstruction. The network has a down-sampling decomposition path (green arrows) and a symmetric up-sampling expansion path (red arrows). The blue arrows mark the paths that skip through the convolutional layers, forming residual blocks. The numbers in italic represent the number of channels in these blocks at different levels. ASP: angular spectrum propagation.

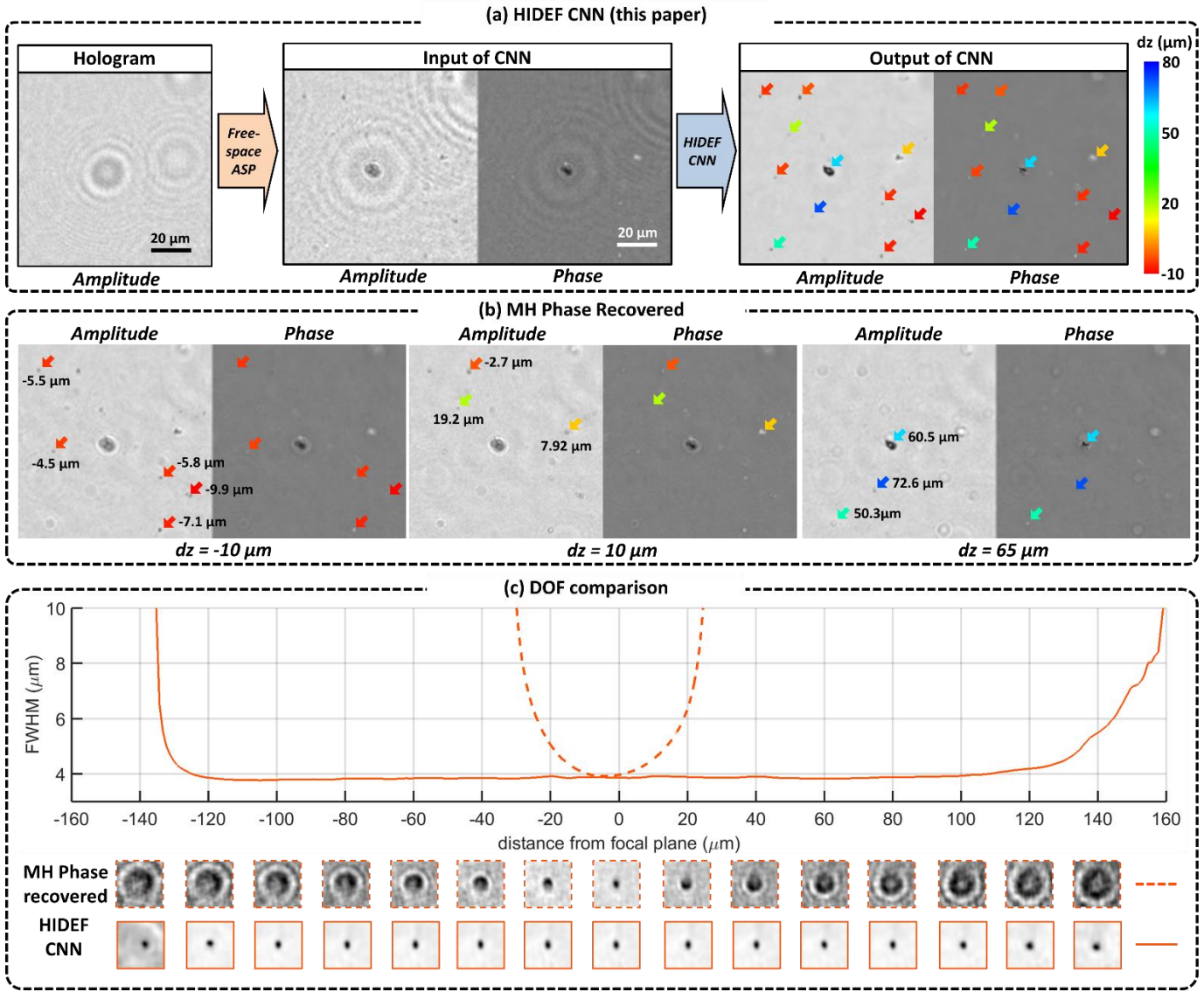

**Fig. 2**. Extended-DOF reconstruction of aerosols at different depths using HIDEF. (a) After its training, HIDEF CNN brings all the particles within the FOV into focus, while also performing phase-recovery. Each particle's depth is color-coded, with respect to the back-propagation distance (1 mm), as shown with the color-bar on the right. (b) As a comparison, MH-PR images of the same FOV show that some of the particles come into focus at different depths, and become invisible or distorted at other depths. For each particle's arrow, the same color-coding is used as in (a). (c) The enhanced-DOF of HIDEF is illustrated by tracking a particle's amplitude full-width-half-maximum (FWHM) as a function of the axial defocus distance (see Supplementary for details). HIDEF preserves the particle's FWHM diameter and its correct image across a large DOF of >0.2 mm, which is expected since it was trained for this range of defocus (± 0.1 mm). On the other hand, MH-PR results show a much more limited DOF, as also confirmed with the same particle's amplitude images at different defocus distances, reported at the bottom.

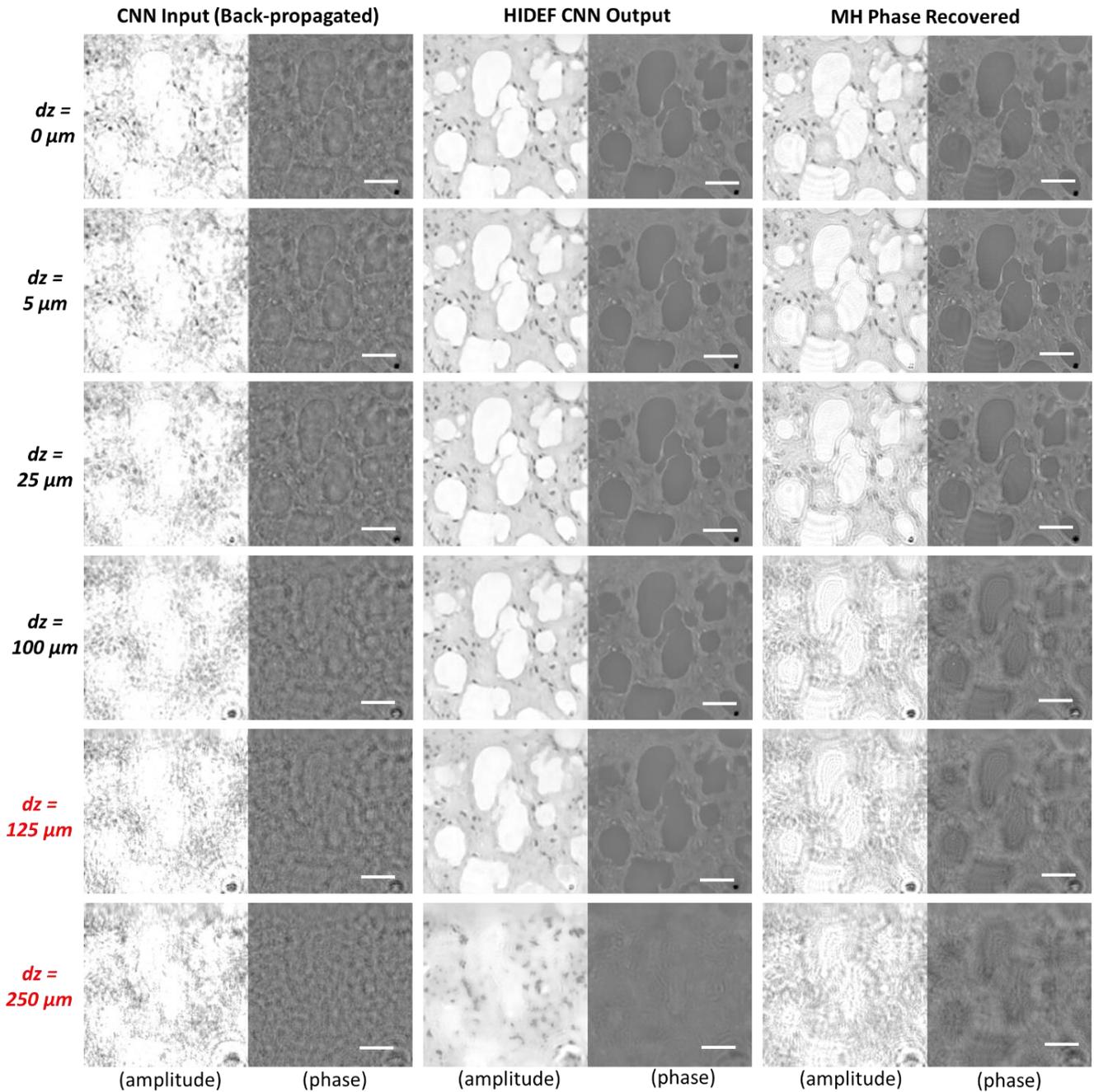

**Fig. 3**. Comparison of HIDEF results against free-space back-propagation (CNN Input) and MH-PR (MH Phase Recovered) results, as a function of axial defocus distance (dz). The test sample is a thin section of a human breast tissue sample. The first two columns use a single intensity hologram, whereas the third column (MH-PR) uses eight in-line holograms of the same sample, acquired at different heights. These results clearly demonstrate that the HIDEF network simultaneously performs phase-recovery and auto-focusing over the axial defocus range that it was trained for (i.e., dz| ≤ 100 μm in this case). Outside this training range (marked with red dz values), the network output is not reliable. Scale bar: 20 μm.

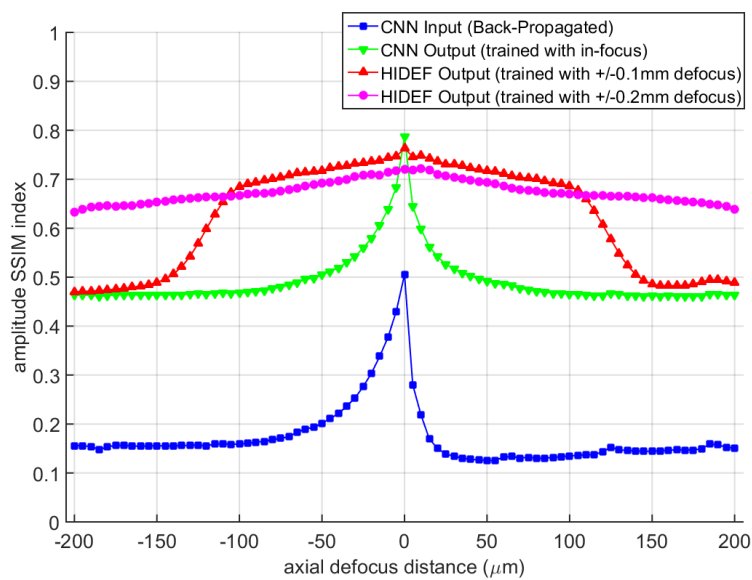

**Fig. 4**. SSIM values as a function of the axial defocus distance. Each one of these SSIM curves is averaged over 180 test FOVs (512-by-512 pixels) corresponding to thin sections of a human breast tissue sample. The results confirm the extended-DOF of the HIDEF network output images, up to the axial defocus range that it was trained for.